\documentclass[sigconf]{acmart} % anonymous
\AtBeginDocument{%
  \providecommand\BibTeX{{%
    \normalfont B\kern-0.5em{\scshape i\kern-0.25em b}\kern-0.8em\TeX}}}

\usepackage{xspace}
\usepackage{algorithm}
\usepackage{algorithmic}
\usepackage{bbding}
\usepackage{multirow}
\usepackage{enumitem}
\usepackage{makecell}
\newcommand{\ie}{{\emph{i.e.}},\xspace}
\newcommand{\eg}{{\emph{e.g.}},\xspace}

\newcommand{\dyf}[1]{\textcolor{black}{#1}}
\copyrightyear{2022}
\acmYear{2022}
\setcopyright{acmcopyright}\acmConference[MM '22]{Proceedings of the 30th ACM International Conference on Multimedia}{October 10--14, 2022}{Lisboa, Portugal}
\acmBooktitle{Proceedings of the 30th ACM International Conference on Multimedia (MM '22), October 10--14, 2022, Lisboa, Portugal}
\acmPrice{15.00}
\acmDOI{10.1145/3503161.3547826}
\acmISBN{978-1-4503-9203-7/22/10}

\acmSubmissionID{356}

\begin{document}

\title{Towards Accurate Post-Training Quantization\\for Vision Transformer}

\author{Yifu Ding$^{1,2,3}$, Haotong Qin$^{1,3}$, Qinghua Yan$^{1}$, \\ Zhenhua Chai$^{2}$,  Junjie Liu$^{2}$, Xiaolin Wei$^{2}$, Xianglong Liu$^{\dagger,1}$ }
\affiliation{\institution{
$^{1}$ State Key Laboratory of Software Development Environment, Beihang University}
\country{}
\state{}
\city{}}
\affiliation{\institution{
$^{2}$ Meituan \quad
$^{3}$ Shen Yuan Honors College, Beihang University}
\country{}
\state{}
\city{}
}

\email{ {yifuding, haotongqin, yanqh, xlliu}@buaa.edu.cn, {chaizhenhua, liujunjie10, weixiaolin02}@meituan.com}

\thanks{$^\dagger$Corresponding author.}

\renewcommand{\shortauthors}{Yifu Ding et al.}

\begin{abstract}
%Post-training quantization is one of the most prevalent approaches to quickly compressing deep networks. Recently, the vision transformer emerges as a potential architecture for computer vision with various varieties springing up. 
%However, existing post-training quantization methods still cause severe performance drop, especially under the ultra-low bit-width settings. 
Vision transformer emerges as a potential architecture for vision tasks. However, the intense computation and non-negligible delay hinder its application in the real world. As a widespread model compression technique, existing post-training quantization methods still cause severe performance drops.
We find the main reasons lie in (1) the existing calibration metric is inaccurate in measuring the quantization influence for extremely low-bit representation, and (2) the existing quantization paradigm is unfriendly to the power-law distribution of Softmax. 
Based on these observations, we propose a novel \textbf{A}ccurate \textbf{P}ost-training \textbf{Q}uantization framework for \textbf{Vi}sion \textbf{T}ransformer, namely \textbf{APQ-ViT}. 
We first present a unified \textit{Bottom-elimination Blockwise Calibration} scheme to optimize the calibration metric to perceive the overall quantization disturbance in a blockwise manner and prioritize the crucial quantization errors that influence more on the final output. 
Then, we design a \textit{Matthew-effect Preserving Quantization} for Softmax to maintain the power-law character and keep the function of the attention mechanism. 
Comprehensive experiments on large-scale classification and detection datasets demonstrate that our APQ-ViT surpasses the existing post-training quantization methods by convincing margins, especially in lower bit-width settings (\eg averagely up to 5.17\% improvement for classification and 24.43\% for detection on W4A4).
We also highlight that APQ-ViT enjoys versatility and works well on diverse transformer variants.
\end{abstract}

%  <concept>\textbf{}

% \ccsdesc[500]{Computing methodologies~Model quantization}

\begin{CCSXML}
<ccs2012>
<concept>
<concept_id>10010147.10010178.10010224.10010245</concept_id>
<concept_desc>Computing methodologies~Computer vision problems</concept_desc>
<concept_significance>500</concept_significance>
</concept>
</ccs2012>
\end{CCSXML}

\ccsdesc[500]{Computing methodologies~Computer vision problems}

\keywords{post-training quantization, vision transformer, computer vision}

\begin{comment}
\begin{teaserfigure}
  \includegraphics[width=\textwidth]{sampleteaser}
  \caption{Seattle Mariners at Spring Training, 2010.}
  \Description{Enjoying the baseball game from the third-base
  seats. Ichiro Suzuki preparing to bat.}
  \label{fig:teaser}
\end{teaserfigure}
\end{comment}
\maketitle

\section{Introduction}

With the development of deep learning, the neural networks achieve great success in a various domains, such as image classification~\cite{krizhevsky2012imagenet,VeryDeepConvolutional,7298594,wang2019dynamic,Wang_2019_ICCV,wang2021dual}, object detection~\cite{DBLP:journals/corr/GirshickDDM13,DBLP:journals/corr/Girshick15,DBLP:journals/corr/abs-1904-02701,NIPS2015_5638,Li_2019_CVPR,WeiOccluded2020}, semantic segmentation~\cite{Everingham:2010:PVO:1747084.1747104,Zhuang_2019_CVPR}, \textit{etc}.
Recently, the Vision Transformer (ViT)~\cite{dosovitskiy2020vit} emerges as a novel and effective architecture and shows great potential for various vision tasks.
However, pretrained models usually have massive parameters and considerable computational overheads, \textit{e.g.}, the ViT-L model is with 307M parameters and 190.7 GFLOPs during inference~\cite{dosovitskiy2020vit}. 
The high computational complexity and non-negligible latency hinder the practical applications of vision transformers in real-world applications especially on edge devices.
To address the challenge, many architectures have been proposed for lightweight vision transformers (\cite{mehta2021mobilevit}, \cite{graham2021levit}, ~\cite{mehta2020delight}). 
Although these works have achieved remarkable speedup and memory footprint reduction, they still rely on floating-point operations, leaving room for further compression by parameter quantization. 

As a model compression approach, quantization compacts the floating-point parameters of neural networks to lower-bit representations, and the computation can be implemented by efficient integer operations on hardware.
Thus, quantized vision transformers significantly save the storage and speed up inference. 
Considering that re-training the transformer is time-consuming and computationally intensive, Post-Training Quantization (PTQ) is a practical solution in widespread scenarios, which just takes a small unlabeled dataset to quantize (calibrate) a pre-trained network with no need for training or fine-tuning.
Many previous works are devoted to quantizing vision transformers~\cite{liu2021post, yuan2021ptq4vit, lin2021fqvit}, which shows great potential in both accuracy and efficiency. Applying the existing methods can almost retain the original accuracy of full-precision transformers under the 8-bit setting.

\begin{figure*}[t]
  \centering
%   \vspace{-0.4in}
  \includegraphics[width=1\linewidth]{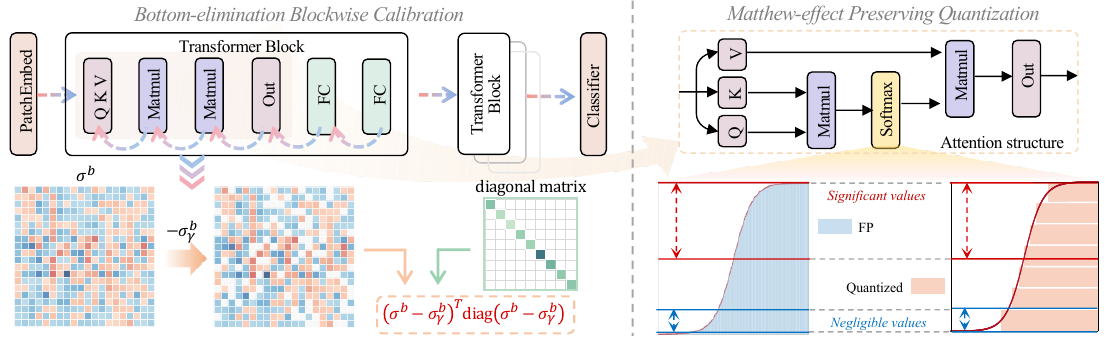}
%   \vspace{-0.25in}
  \caption{Overview of APQ-ViT. The left is Bottom-elimination Blockwise Calibration to apply quantization in a blockwise manner to perceive the quantization loss of adjacent layers, and prioritize the significant errors by eliminating the second-order gradient corresponding to trivial errors. The right is Matthew-effect Preserving Quantization, which is specialized for maintaining the power-law distribution of the Softmax function. }
  \label{fig:overview}
\end{figure*}

However, when quantizing the vision transformer to ultra-low bit-widths (\eg, 4-bit weight and activation), the model suffers severe accuracy drop or even crashes.
Our study reveals that the poor performance might attribute to two issues from optimization and structure perspectives. 
From the optimization perspective, the limited bit-width constraints the representation capability and causes larger errors which makes the existing second-order layerwise calibration metric not accurate in measuring the impact of quantization error on the final output. % it difficult to maintain the information in the original model after mapping to the fixed-point. 
And as for the structure perspective, the existing quantization paradigm is unfriendly to the Softmax function in the attention mechanism, which is also known as a normalized exponential function. It redistributes the inputs to satisfy the power-law probability, while we discover that previous quantization solutions are easy to damage the Matthew-effect of Softmax. 
Therefore, specializing in the quantization strategy for vision transformers is a great need to improve the accuracy of the low-bit quantized model. 

In this paper, we propose an accurate post-training quantization method for vision transformers, namely APQ-ViT, which considers both the optimization difficulty and the special structure for low bit-width (See the overview in Figure~\ref{fig:overview}).
First, we present a unified \textit{Blockwise Bottom-elimination Calibration} (BBC) scheme to optimize the calibration metric based on the block-stacking architecture, which can be flexibly generalized to other variants. It enables the metric to perceive the quantization loss in a blockwise manner and prioritize the significant errors by eliding the second-order gradients corresponding to the inevitable trivial errors. 
Second, the \textit{Matthew-effect Preserving Quantization} (MPQ) is specifically tailored for the Softmax function. Instead of obeying the maximizing mutual information paradigm as many quantization methods, we hold the view that preserving the power-law distribution in the quantized Softmax is more crucial for the attention mechanism.

Our APQ-ViT revisits the process of post-training quantization for vision transformer and presents novel insights.
Comprehensive experiments on the large-scale computer vision tasks (image classification~\cite{Deng2009ImageNet} and object detection~\cite{lin2014coco}) demonstrate that our APQ-ViT performs remarkably well across various transformer architectures such as ViT~\cite{dosovitskiy2020vit}, DeiT~\cite{Touvron2021deit}, and Swin Transformer~\cite{ze2021swin}, and surpasses the existing methods by convincing margins, especially in lower bit-width settings (\eg averagely up to 5.17\% improvement for classification and 24.43\% for detection on W4A4).
We highlight that our APQ-ViT scheme achieves state-of-the-art accuracy performance on various bit-width settings, and enjoys versatility on diverse architectures and vision tasks. 

We summarize our main contributions as: 
\begin{itemize}[nosep, leftmargin=*] 
    \item We find that for the post-training quantization of vision transformer, (1) the extremely low-bit representation makes the existing calibration metric inaccurate in measuring the quantization errors; and (2) an inconsistency exists between the quantization paradigm and the power-law distribution of Softmax.  %an inconsistency exists between mutual-information-based quantization principle and the power-law redistribution of Softmax function. 
    \item We present an accurate post-training quantization method for vision transformer, namely APQ-ViT, with a unified Blockwise Bottom-elimination Calibration scheme to enable the quantization perception inside blocks and prioritize the crucial errors that influence the final predictions. 
    \item Our study reveals the power-law distribution of the Softmax function and proposes the Matthew-effect Preserving Quantization. In contrast to purely minimizing the quantization loss, it inspires a novel perspective to preserve the character of Softmax while embedding quantization. % these insights might help to inspire how to embed quantization rationally in attention mechanisms. 
    \item We evaluate the APQ-ViT on large-scale image classification and object detection tasks with different model variants and bit-width, and obtain prevailing improvements over existing post-training quantization methods especially in lower bit-width. 
\end{itemize}

\begin{figure*}[t]
  \centering
%   \vspace{-0.4in}
  \includegraphics[width=0.95\linewidth]{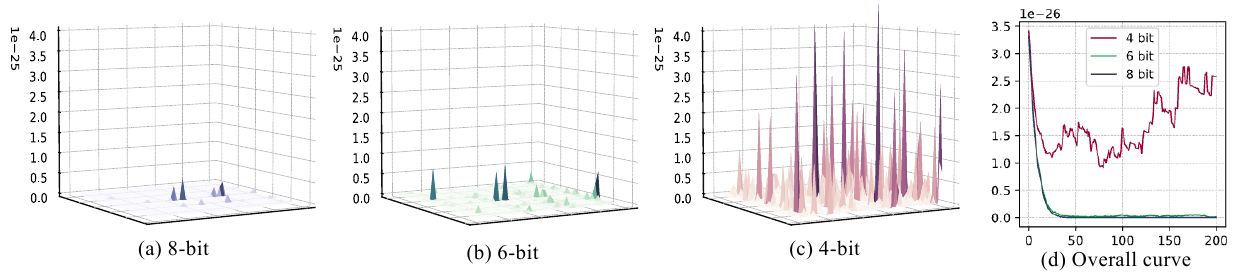}
%   \vspace{-0.25in}
  \caption{Visualization of Hessian-guided loss term of the optimal scaling factor for (a) 8-bit, (b) 6-bit and (c) 4-bit quantization. And (d) shows the curve of overall loss terms for all the candidates. }
  \label{fig:3d-hessian}
\end{figure*}

\section{Related Work}

\subsection{Vision Transformer}

%The vision transformer shows a great potential in many vision tasks, including classification, object detection, semantic segmentation, and video understanding. 
%The classical vision transformer is constructed pure transformer targeting to process image patches and use multi-head self-attention as a sequence. 
%Specifically, a vision transformer is stacked by several attention-based blocks, where each block is composed of a multi-head self-attention module (MSA) and multi-layer perceptron (MLP). MSA generates the attention between different patches to extract features with global information. Typical MLP contains two fully-connected layers (FC) and the GELU activation function is used after the first layer. The input sequence is first fed into each self-attention head of MSA. In each head, the sequence is linearly projected to three matrices, query $Q=X W^{Q}$, key $K=X W^{K}$, and value $V=X W^{V}$. 
%Then, matrix multiplication $Q K^{T}$ calculates the attention scores between patches. The Softmax function is used to normalize these scores to attention probability $P$. The output of the head is matrix multiplication $P V$. 
%The process is formulated as Eq.~\ref{eq:attention}:
%\begin{equation}
%\label{eq:attention}
%\operatorname{Attention}(Q, K, V)=\operatorname{softmax}\left(\frac{Q K^{T}}{\sqrt{d}}\right) V
%\end{equation}
%where $d$ is the hidden size of head. The outputs of multiple heads are concatenated together as the output of MSA.
The classical vision transformer~\cite{dosovitskiy2020vit} is constructed by pure transformer targeting to process image patches, which is stacked by several attention-based blocks that are composed of a multi-head self-attention module and multi-layer perceptron.
DETR~\cite{carion2020detr} further extends to object detection, which uses ResNet as the backbone and replaces the detection head with transformers. Following them, there are many variants for more applications with new techniques specialized for CV applications~\cite{ranftl2021vision, prangemeier2020attention, esser2021taming, kumar2021colorization}. 
Swin Transformer~\cite{ze2021swin} emerges as a competitive backbone with excellent generalization capability for many benchmarks and remarkably surpasses the state-of-the-art CNNs in most CV tasks. 

Many works are also devoted to balancing performance and efficiency. MobileViT exploits global representation capability and spatial order preservation of transformer by inserting it into convolution blocks. Some works simplify the attention mechanism, like sparse attention~\cite{Beltagy2020Longformer, zaheer2020big}, linear approximate attention~\cite{katharopoulos2020transformers, choromanski2020rethinking} QuadTree attention~\cite{tang2022quadtree}, or hashing-based attention~\cite{kitaev2020reformer}. Moreover, general model compression techniques are also actively applied. DeiT introduces distillation tokens to better interact with the teacher model through attention. \cite{zhu2021visual} prunes vision transformers by sparsing the unnecessary feature channels by ranking the importance. \cite{he2021spvit} combines NAS and parameter sharing to search and replace some self-attention modules by convolutions to improve the locality extraction. The above methods focus on optimizing the transformer architectures while keep the parameters full-precision, leaving room for further compression by quantization. 

\subsection{Quantization}

Model quantization is one of the promising compression approaches, which quantizes the full-precision parameters to lower bit-width. It can not only shrink the model size but reduce computational complexity by transforming floating-point calculations to fixed-point, which significantly accelerates the inference, decreases the memory footprint, and reduces energy consumption. 
Current quantization methods can be categorized as Quantization-aware Training (QAT) and Post-training Quantization (PTQ) according to whether training/fine-tuning or not. Considering that training vision transformers is computationally intensive and time-consuming, they always have a huge demand for computation and power resources that emits lots of carbon footprint. PTQ, a training-free method, is well recognized as a more feasible solution, which can be broadly divided into two types: 1) searching best scale factor. 2) optimizing calibration strategy. To search best scale factors, \cite{choukroun2019omse} proposes an optimal MSE to select the scale factor that minimizes the quantization error. \cite{yuan2021ptq4vit} uses Twin Uniform Quantization that specially designs two scales for long-tail parameter distribution of Softmax and GeLU, and proposes Hessian Guided Metric to search for best scales. ~\cite{fang2020post} also use Piecewise Linear Quantization to make the quantized parameters better fit the bell-shaped distribution of weight and activation after scaling. As for the calibration strategy,  AdaQuant~\cite{jhunjhunwala2021adaquant, hubara2021accurate} utilizes layerwise optimization, which fixes the error induced by quantizing former layers by sequential calibration. EasyQuant~\cite{easyquant} uses an alternative scale optimization of weight and activation, fixing one and optimizing the other throughout the network. \cite{bai2021towards} minimizes the quantization error by module-wise reconstruction to jointly optimize all the coupled linear layers inside each module.

\section{Method}

% In this section, we first revisit the quantization error distribution the post-training quantization of vision transformers, and we analysis the information amount of applying different scaling factors by diffirential entropy. We present our Information Distribution Guided Metric and Headwise Adaptable Log Quantization according to the information pattern that obeys minimization of error quantization. Finally, we provide an alternative optimization strategy for better convergence. 
%In this section, we first build a post-training quantization scheme for vision transformers, and then theoretically analyze the different quantization solution for the Softmax function. Moreover, we propose a magnitude-aware error minimization with importance sampling which focus more on larger errors. The overview framework of our scheme is Figure~\ref{}.
In this section, we propose an accurate post-training quantization framework for vision transformer, namely APQ-ViT. 
We first present the basic quantization pipeline and then introduce our techniques, including 
\textit{Blockwise Bottom-elimination Calibration} (BBC) to tackle the optimization difficulties in low bit-width and the \textit{Matthew-effect Preserving Quantization} (MPQ) to preserve the power-law redistribution for Softmax function. 

\subsection{Preliminaries}
\label{sec:pre}
% Among the existing post-training quantization methods for vision transformer, one of representative is layerwise calibration by Hessian-guided metric~\cite{yuan2021ptq4vit}.

As a widespread solution, the asymmetric uniform quantization is applied to quantize network.
And in a standard quantization transformer, the input data first passes through a quantized embedding layer before being fed into the quantized transformer blocks, and each transformer block consists of an MSA module and an MLP.
%\textbf{Multi-head attention.}
The computation of MSA depends on queries $\mathbf{Q}$, keys $\mathbf{K}$ and values $\mathbf{V}$, which are derived from hidden states $\mathbf{H}$. In a specific quantized transformer layer, $\mathbf{H}$ is first quantized to $\hat{\mathbf{H}}$ before passing through linear layers, which can be expressed as
% the computation in an attention head can be expressed as
\begin{equation}
\hat{\mathbf{Q}}=\hat{\mathbf{w}}_Q \ \hat{\mathbf{H}}, \  
\hat{\mathbf{K}}=\hat{\mathbf{w}}_K \ \hat{\mathbf{H}}, \  
\hat{\mathbf{V}}=\hat{\mathbf{w}}_V \ \hat{\mathbf{H}},
\end{equation}
% where ${\operatorname{linear}_k^Q}, {\operatorname{linear}_k^K}, {\operatorname{linear}_k^V}$ represent three different $k$-bit quantized linear layers for $\mathbf{Q},\mathbf{K},\mathbf{V}$ respectively. 
where $\hat{\mathbf{w}_Q}$, $\hat{\mathbf{w}_K}$, $\hat{\mathbf{w}_V}$ represent quantized weight of three different linear layers for $\mathbf{Q}$, $\mathbf{K}$, $\mathbf{V}$ respectively. 
The computation of self-attention is formulated as Eq.~(\ref{eq:attention}):
\begin{equation}
\label{eq:attention}
\operatorname{Attention}_q(\mathbf Q, \mathbf K, \mathbf V)= \operatorname{softmax}_q\left(\frac{\hat{\mathbf Q} \times \hat{\mathbf K}^{T}}{\sqrt{d}}\right)  \times \hat{\mathbf V},
\end{equation}
where $d$ is the hidden size of the head and $\operatorname{softmax}_q$ denotes the Softmax function with quantized output. The outputs of multiple heads are concatenated together as the output of MSA.
Moreover, the MLP contains two quantized linear layers, and the GeLU activation function is used after the first layer.

Among the existing post-training quantization methods for vision transformers, one of the representatives is PTQ4ViT~\cite{yuan2021ptq4vit}, which is a typical representation of these works using the Hessian guided metric to determine the scaling factors. 
In classification task, the task loss is $\mathcal{L}=\mathrm{CE}(\hat{y}, y)$, where CE is cross-entropy, $\hat{y}$ is the output of the network and $y$ is the ground truth. 
The expectation of loss is a function of network parameters $\mathbf{x}$, which is $\mathbb{E}[\mathcal{L}(\mathbf{x})]$. The quantization brings a small perturbation $\epsilon$ on parameter $\hat{\mathbf{x}}=\mathbf{x}+\epsilon$. We analyze the influence of quantization to the task loss by Taylor series expansion:
\begin{equation}
\label{eq:taylor}
\mathbb{E}[\mathcal{L}(\hat{\mathbf{x}})]-\mathbb{E}[\mathcal{L}(\mathbf{x})] \approx \epsilon^{T} \bar{g}^{(\mathbf{x})}+\frac{1}{2} \epsilon^{T} \bar{H}^{(\mathbf{x})} \epsilon,
\end{equation}
where $\bar{g}^{(\mathbf{x})}$ is the gradients and $\bar{H}^{(\mathbf{x})}$ is the Hessian matrix. The target is to find the scaling factors to minimize the influence: $\min _{\Delta}(\mathbb{E}[\mathcal{L}(\hat{\mathbf{x}})]-\mathbb{E}[\mathcal{L}(\mathbf{x})])$.
%, and then follow the layer-wise reconstruction method in \cite{}.
Since weight's perturbation $\epsilon$ is relatively small, we have a first-order Taylor expansion that $(\hat{O}-O) \approx g_{O}(\mathbf{x}) \epsilon$, where $\hat{O}=(\hat{\mathbf{x}}+\epsilon)^{T} \hat{\mathbf{x}}$. The second-order term in Eq.~(\ref{eq:taylor}) could be written as
\begin{equation}
% \begin{aligned}
% \mathcal{L}=\epsilon^{T} \bar{H}^{(W)} \epsilon &=\left(J_{O}(W) \epsilon\right)^{T} \bar{H}^{(O)} J_{O}(W) \epsilon \\
% & \approx(\hat{O}-O)^{T} \bar{H}^{(O)}(\hat{O}-O).
\epsilon^{T} \bar{H}^{(\mathbf{x})} \epsilon \approx(\hat{O}-O)^{T} \bar{H}^{(O)}(\hat{O}-O).
% \end{aligned}
\end{equation}
%Based on the layer-wise reconstruction method in \cite{}, the optimization can be approximated by:
%\begin{equation}
%\min _{\Delta} \mathbb{E}\left[\left(\hat{O}^{l}-O^{l}\right)^{T} \operatorname{diag}\left(\left(\frac{\partial L}{\partial O_{1}^{l}}\right)^{2}, \ldots,\left(\frac{\partial L}{\partial O_{\left|O^{l}\right|}^{l}}\right)^{2}\right)\left(\hat{O}^{l}-O^{l}\right)\right] 
%\end{equation}
%where $O^{l}$ and $\hat{O}^{l}$ are the outputs of the $l$-th layer before and after quantization, respectively.
Then we follow \cite{easyquant, yuan2021ptq4vit} to traverse the search spaces of $\Delta_{\mathbf{x}}$ by linearly dividing $\left[\alpha \frac{\mathbf{x}_{\max }-\mathbf{x}_{\min }}{2^{k}}, \beta \frac{\mathbf{x}_{\max }-\mathbf{x}_{\min }}{2^{k}}\right]$ to $n$ candidates. $\alpha$ and $\beta$ are two parameters to control the search range. We alternatively search for the optimal scaling factors $\Delta_{\mathbf{w}}^{*}$ and $\Delta_{\mathbf{\mathbf{a}}}^{*}$ in the search space. Firstly, $\Delta_{\mathbf{\mathbf{a}}}$ is fixed, and we search for the optimal $\Delta_{\mathbf{\mathbf{w}}}$ to minimize loss $\mathcal{L}$. Secondly, $\Delta_{\mathbf{\mathbf{w}}}$ is fixed, and we search for the optimal $\Delta_{\mathbf{\mathbf{a}}}$ to minimize $\mathcal{L}$. $\Delta_{\mathbf{w}}$ and $\Delta_{\mathbf{a}}$ are alternately optimized for several rounds.

\subsection{Blockwise Bottom-elimination Calibration}

From the optimization perspective, existing typical post-training quantization methods for vision transformers use the second-order Hessian-guided metric to measure the quantization loss caused by each candidate scaling factor and then determine the optimal quantizer. However, we find that for the extremely low-bit representation, the layerwise optimization is inaccurate since it is unable to perceive the quantization in a higher block scale, and the quantization error is inevitably larger while the dense Hessian matrix loses the attention of the significant errors. 
% However, we found that even applying advanced techniques, the quantization process still faces optimization difficulties.

Ideally, we expect to determine the quantizer with the smallest quantization loss by a carefully designed loss term, and the loss should be significantly smaller compared to other candidates which converges to a local optimum. 
But in practice, we find that the Hessian-guided loss terms calculated by each candidate have large variance, especially at ultra-low bits (such as 4-bit). As shown in Figure~\ref{fig:3d-hessian}, when we quantize the model to 8-bit, the quantization loss is steady and relatively small, the optimal loss plane in Figure~\ref{fig:3d-hessian}(a) is also flat. 
However, when the bit-width is reduced to 4-bit, the behavior of adjacent candidates shows a significant difference, and the loss curve fluctuations heavily. Even the optimal candidate has lots of spikes on the loss plane (see Figure~\ref{fig:3d-hessian}(c)). 
The phenomena reveal the defects of the current calibration strategy, due to the higher degree of discretization in lower-bit quantization, (1) the loss in a single layer is larger, which has an impact on the calibration of other layers in the layerwise calibration strategy, (2) the quantization loss of each element varies greatly that times larger than 8-/6-bit quantization. 
% When we reduce the quantization bit-width, a higher degree of discretization brings more difficulties in optimization: 

Therefore, we propose a \textbf{B}lockwise \textbf{B}ottom-elimination \textbf{C}alibration (BBC) scheme for the post-training quantization. It optimizes the calibration in a blockwise manner which enables the Hessian-guided loss to have a perception of the quantization error of adjacent layers in a single block. And It uses the bottom-elimination mechanism to focus on the critical errors that influence the final output instead of the whole loss plane. 

Firstly, we built a blockwise calibration scheme from post-training quantization.
Taking $b$-th block with $L$ layers as an example, the computation in the block can be represented as
\begin{equation}
O^{b}={\mathbf{w}_L^b}^T {\mathbf{w}_{L-1}^b}^T\cdots {\mathbf{w}_1^b}^T \mathbf{a}^b,
\end{equation}
where $\mathbf{a}^b$ and $O^{b}$ denote the input and output of the $b$-th transformer block.
When the $l$-th layer is calibrated, the $L$-th to $l$-th layers can be considered as a composite layer, and its weight and activation is expressed as 
\begin{equation}
\mathcal{W}_l^b={\mathbf{w}_L^b}^T\cdots {\mathbf{w}_{l+1}^b}^T {\mathbf{w}_l^b}^T, \quad
\mathcal{A}_l^b={\mathbf{w}_{l-1}^b}^T\cdots {\mathbf{w}_1^b}^T \mathbf{a}^b,
\end{equation}
% The quantized weight $\hat{\mathcal{W}}_l$ of the composite layer is
% \begin{equation}
% \hat{\mathcal{W}_l^b}=\hat{\mathbf{w}_L^b}^T\cdots \hat{\mathbf{w}_{l+1}^b}^T \hat{\mathbf{w}_l^b}^T.
% \end{equation}
Taking the weight calibration as an example, the second-order term of the $l$-th layer in $b$-th transformer block can be expressed as
\begin{equation}
\label{eq:loss_1}
\begin{aligned}
{\epsilon_l^b}^{T} \bar{H}^{(\mathcal{W})} {\epsilon_l^b}
=&\left(J_{O^b}(\mathcal{W}_l^b) {\epsilon_l^b}\right)^{T} \bar{H}^{(O^b)} J_{O^b}(\mathcal{W}_l^b) {\epsilon_l^b} \\
\approx&\mathbb{E}\left[{\sigma^b}^{T} \operatorname{diag}\left(\left(\frac{\partial \mathcal L}{\partial O_{1}^{b}}\right)^{2}, \ldots,\left(\frac{\partial \mathcal L}{\partial O_{\left| O^b \right|}^{b}}\right)^{2}\right){\sigma^b}\right],
\end{aligned}
\end{equation}
where $\epsilon_l^b=\mathcal{W}_l^b-\hat{\mathcal{W}_l^b}$, $\sigma^b=\hat{O}^{b}-O^{b}$, and $O^{b}$, $\hat{O}^{b}$ are the outputs of the $b$-th block before and after quantization, respectively. $O^b_i$ denotes the $i$-th dimension of $O^b$, where $i \in [1,|O^b|]$. 
% Therefore, we select the candidate for each layer in a blockwise manner by reducing the impact on the final outputs. 
Therefore, we optimize the calibration metric to enable the perception of the whole block and reduce the impact on the final output. 

\begin{figure}
  \centering
  \includegraphics[width=3.0in]{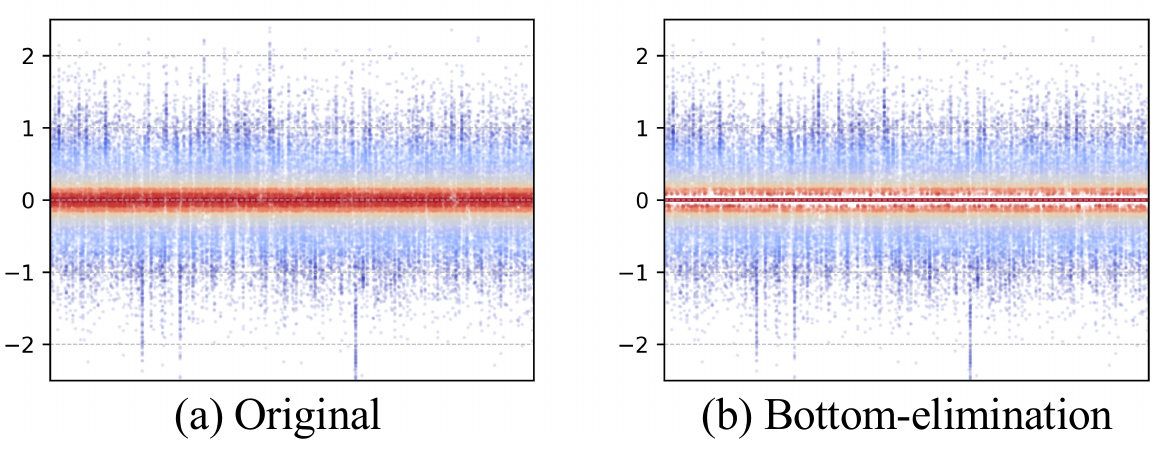}
  \caption{The quantization error measured by Hessian-based metric. Y-axis represents the magnitude of errors. (a) is the whole error distribution, and (b) visualizes the elimination of error in the 10th percentile. }
  \label{fig:bottom-elimination}
\end{figure}

Secondly, since the quantization error is deemed as inevitable in rounding operation and significantly increases when the bit-width is smaller, the larger errors should be of major concern. Inspired by \cite{ye2020accelerating, zhang2020memorized, sun2017meprop, aji2017sparse} that pruning the gradients close to zero in the backward propagation has a tiny impact on weight-updating. 
To pay more attention to the large errors that perturb the final output and also strike a balance between the range and variance of error, we further propose the bottom-elimination mechanism to obtain a sparse Hessian matrix in the optimization scheme. It considers the Hessian-based metric as weighted second-order gradients, and prunes the gradients that correspond to the smallest absolute quantization errors. 
Specifically, we construct a bottom-elimination matrix $\sigma^b_\gamma$ for the $\sigma^b=\hat{O^b}-O^b$, which aims to select the elements with absolute values in $\gamma$-th percentile:
\begin{equation}
\sigma^b_\gamma=\sigma^b[\sigma^b]_{\gamma}, \qquad
[\sigma^b]_{\gamma}=
\left\{
\begin{aligned}
1, & \text{ where } \sigma^b < |\sigma^b|_\gamma,\\
0, & \text{ otherwise},
\end{aligned}
\right.
\end{equation}
where $[\cdot]$ denotes the \textit{Iverson bracket}~\cite{iverson1962programming}. 
We apply the bottom elimination matrix $\sigma^b_\gamma$ to Eq.~(\ref{eq:loss_1}) so that the obtained metric reflects the critical elements in quantization that causes larger quantization error, and the optimization objective function is expressed as:
\begin{equation}
\label{eq:loss_bbc}
\min_{\Delta}\mathbb{E}\left[\left({\sigma^b-\sigma^b_\gamma}\right)^{T} \operatorname{diag}\left(\left(\frac{\partial \mathcal L}{\partial O_{1}^{b}}\right)^{2}, \ldots,\left(\frac{\partial \mathcal L}{\partial O_{\left| O^b \right|}^{b}}\right)^{2}\right)\left({\sigma^b-\sigma^b_\gamma}\right)\right].
\end{equation}
Figure~\ref{fig:bottom-elimination} visualizes the effect of the bottom-elimination mechanism for quantization errors. With the matrix $\sigma_\gamma$, we optimize the calibration metric to make it focuses on perturbations with large magnitude which have nonnegligible influence on the final output of task predictions.

\begin{figure}
  \centering
  %  \vspace{-0.4in}
  \includegraphics[width=3.5in]{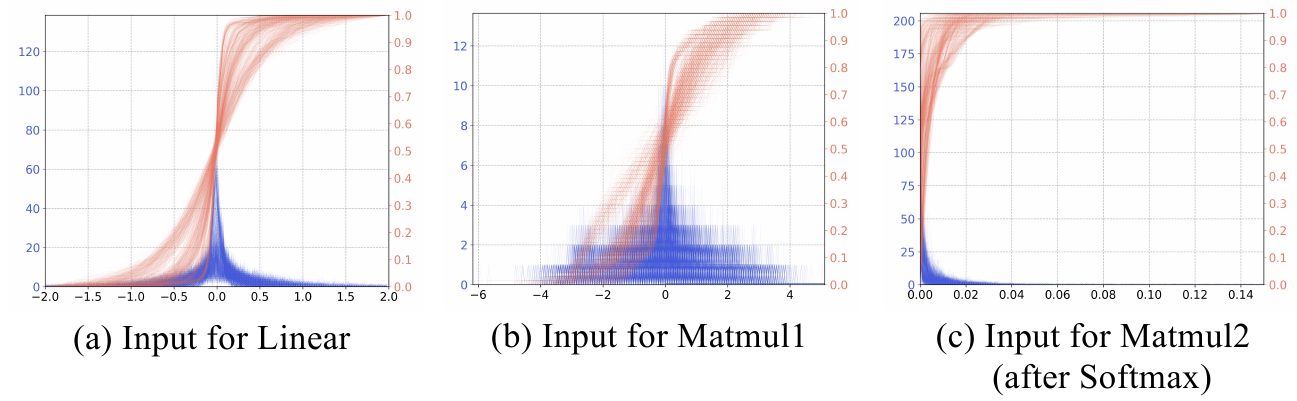}
  %  \vspace{-0.25in}
  \caption{Visualization of different input activation distribution in the pretrained vision transformer model. (c) presents an extreme unbalanced distribution. }
  \label{fig:input-activation}
\end{figure}

\subsection{Matthew-effect Preserving Quantization}

The special architecture of the attention mechanism in the vision transformer is also an obstacle to low-bit quantization. Especially the Softmax function, also known as the normalized exponential function, is well recognized to be unfriendly to quantization. Generally speaking, as shown in Section~\ref{sec:pre}, each block in the vision transformer usually contains three types of computation: Linear operation, Matmul operation, and Softmax operation, these three operations involve the majority of quantized representations. We show a typical distribution of the activation inputs in Figure~\ref{fig:input-activation}. We can see that the input distributions of the Linear and Matmul1 are similar to the Gaussian and Laplacian distributions that common methods can well quantize. However, the output of the Softmax operation obeys the power-law probability distribution, which is asymmetric and extremely unbalanced.

As a consensus, the ideal quantized parameters should retain the information of full-precision counterparts as much as possible, which is formulated as:
\begin{equation}
\arg\max\limits_{\mathbf{x}, \hat{\mathbf{x}}}\ \mathcal{I}(\mathbf{x} ; \hat{\mathbf{x}}) =\mathcal{H}(\mathbf{x})-\mathcal{H}(\hat{\mathbf{x}} \mid \mathbf{x}),
\end{equation}
where ${\mathcal{H} (\hat{\mathbf{x}})}$ is the information entropy, and ${\mathcal {H} (\hat{\mathbf{x}} \mid \mathbf{x})}$ is the conditional entropy of $\hat{\mathbf{x}}$ given $\mathbf{x}$.
Since we use the deterministic quantization function, the value of $\hat{\mathbf{x}}$ fully depends on the value of $\mathbf{x}$, \textit{i.e.}, $\mathcal{H}(\hat{\mathbf{x}}\mid \mathbf{x})=0$. Thus, the objective function is equivalent to maximizing the information entropy:
\begin{equation}
\arg\max\limits_{\hat{\mathbf{x}}}\ \mathcal{H}(\hat{\mathbf{x}})=-\sum_{\hat{x} \in \hat{\mathcal{X}}} p_{\hat{\mathbf{x}}}(\hat{x}) \log p_{\hat{\mathbf{x}}}(\hat{x}),
\end{equation}
where $p_{\hat{\mathbf{x}}}$ denotes the probability mass function of quantized parameter $\hat{\mathbf{x}}$.
The formulation suggests that a well-optimized quantizer tends to make the probabilities in each quantization interval equal. 

\begin{figure}[t]
  \centering
  %  \vspace{-0.4in}
  \includegraphics[width=2.7in]{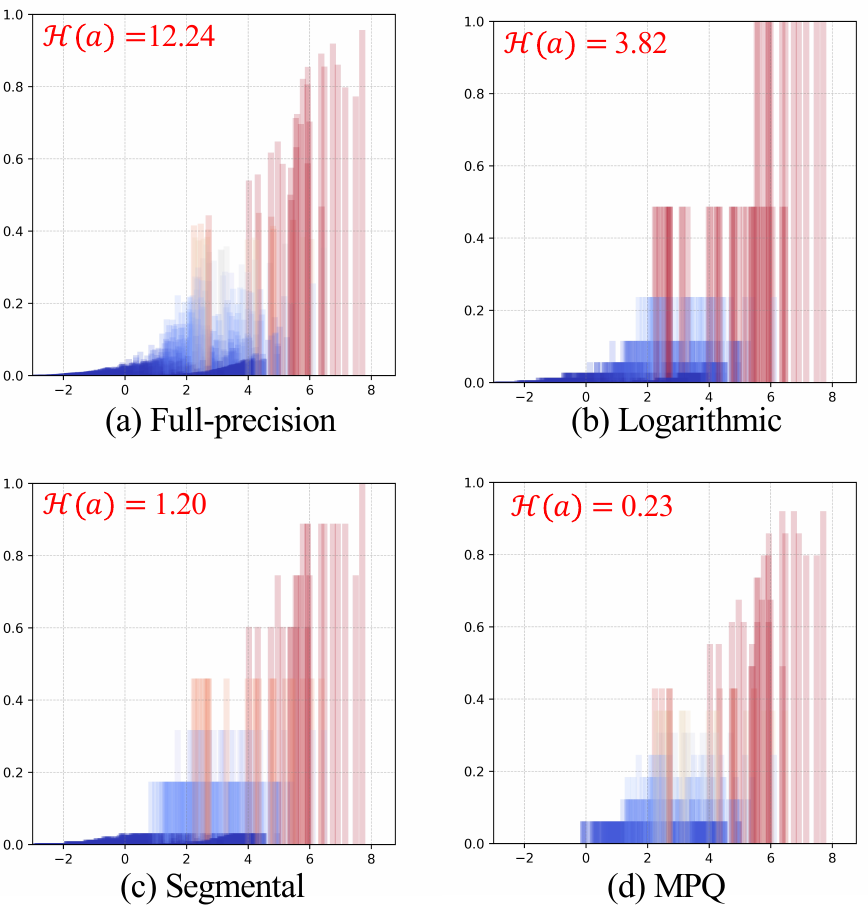}
  %  \vspace{-0.25in}
  \caption{Comparison of quantizing attention scores with different quantizers under 4-bit setting. The x-axis is the attention values before softmax, and the y-axis is (a) full-precision or quantized to 4-bit using (b) Logarithmic quantizer, (c) Segmental quantizer~\cite{yuan2021ptq4vit} and (d) MPQ quantizer. The left-top values are mutual information of each distribution. }
  % The original full-precision scores present an obvious orientation of Matthew-effect with an asymmetric long-tailed dense distribution of small values (the dark blue bars), and the sparse distribution of large values (the red bars). 这一段要加这里吗，caption 有点长
  \label{fig:softmax}
\end{figure}

Interestingly, we observe that the quantization behavior of the Softmax function breaks the widespread idea in a counter-intuitive way. 
Softmax is often used as an activation function to stable the network since it can control the largest value computed in each exponent, which can be written as:
\begin{equation}
\mathrm{softmax}(\mathbf{x})_i = \frac{e^{\beta x_i}}{\Sigma^{K}_{j=1} e^{\beta x_j}}, \mathrm{for}~i = 1, \dots, K,
\end{equation}
where $\beta \in \mathbb{R}$, and usually $\beta = 1$ in neural networks. It creates extreme asymmetric and unbalanced distributions by converting to the exponent. Therefore, many methods are devoted to designing specific quantizers for the quantization of Softmax output to maximize the information, such as Segmental quantizers~\cite{yuan2021ptq4vit, fang2020post}, Logarithmic quantizers~\cite{lin2021fqvit, kim2021bert} or apply sparsification before quantization ~\cite{ji2021onthedistribution}. 
As shown in Figure~\ref{fig:softmax}, the Logarithmic quantizer has the largest 3.82 mutual information. It utilizes up to 11 intervals (about 69\%) to represent the values clustered in $[0, 0.01]$ (about 89\%), while only spares 5 fixed-points to map the rest range $(0.01, 1]$. And Segmental quantizer also utilizes more bits to cover the small values. 

However, consider the original function of Softmax, when $\beta>0$, the function will create probability distributions that are more concentrated around the positions of the largest input values. To put it simply, Softmax makes the small values even smaller while the larger values take the most probability, which is a typical phenomenon of the Matthew-effect. This character helps neural networks to stable the activation flow. Transformer architecture also adopts the Softmax as activation functions to compute the attention scores which measures the relationship of one patch in the sequence with all the other patches. Unfortunately, existing quantizers prioritize the overall mutual information while ignoring the Matthew-effect of the Softmax function. For the significant large values, the Logarithmic and Segmental quantizers only spare fewer bits, thus the information in the larger values is damaged.

Therefore we present a \textbf{M}atthew-effect \textbf{P}reserving \textbf{Q}uantization (MPQ) to quantize the Softmax output. It does not purely pursue the maximization of mutual information before and after quantization, but maintains the Matthew-effect of Softmax output during the quantization process. 
A typical MPQ is an asymmetric linear quantization, which can be expressed as:
\begin{equation}
\begin{aligned}
%\label{eq:quantization}
\hat{\mathbf{x}_s} = \operatorname{clamp} \left(\left \lfloor \frac{\operatorname{softmax}(\mathbf{x})}{\Delta}\right \rceil, 0, 2^k-1\right),\quad
\Delta = \frac{\max(\operatorname{softmax}(\mathbf{x}))}{2^{k}-1}.
\end{aligned}
\end{equation}
where $\Delta$ is the scaling factor, and $\left\lfloor \cdot \right\rceil$ means rounding operation.
MPQ is a straightforward method that brings no extra implementation and inference overhead, and we find that it significantly improves the performance compared with other quantizers. 
As shown in Figure ~\ref{fig:softmax}, MPQ allocates more bits to the larger range, where the values are sparse but significant. In this way, the important values are quantized with finer representations, which better preserves the function of Softmax.

\renewcommand{\algorithmicrequire}{\textbf{Input:}}
\renewcommand{\algorithmicensure}{\textbf{Output:}}
\begin{algorithm}
\caption{The optimization pipeline of APQ-ViT}    
\label{alternative-pipeline} 
\begin{algorithmic}[1]
\REQUIRE Pre-trained vision transformer model and calibration set;
\ENSURE Optimal scaling factors $\mathbf \Delta^{*}$ for all layers;
\FOR{$b=1:\#Block$}
\STATE Forward propagation to get full-precision $O^b$ with the computation process $AB$ in each layer and get loss $\mathcal L$;
\ENDFOR
\FOR{$b=1:\#Block$}
\STATE Backward propagation to get $\frac{\partial \mathcal L}{\partial O^{b}}$;
\ENDFOR
\FOR{$b=1:\#Block$}
\FOR{$l=\#Layer:1$}
\STATE initialize scaling factor $\Delta_{B^{b}_{l}}^{*} \leftarrow \frac{B^{b}_{l \max} - B^{b}_{l \min}}{2^{k}}$ for $B^{b}_l$;
\FOR{search\_round$=1:\#Round$}
\STATE searching scaling factor $\Delta_{A^{b}_l}^{*}$ for $A^{b}_l$ using Eq.~(\ref{eq:loss_bbc}); 
\STATE searching scaling factor $\Delta_{B^{b}_l}^{*}$ for $B^{b}_l$ using Eq.~(\ref{eq:loss_bbc});  
\ENDFOR
\ENDFOR
\ENDFOR
\RETURN Optimal scaling factors $\mathbf \Delta^{*}$.
\end{algorithmic}
\end{algorithm}

\subsection{Framework of APQ-ViT}

We propose an Accurate Post-training Quantization framework for vision transformer, namely APQ-ViT. The quantization process is shown as Algorithm~\ref{alternative-pipeline}.
The APQ-ViT mainly depends on two novel techniques: BBC aims to improve the second-order calibration metric and MPQ specializes in the Softmax structure. 
In the calibration process, APQ-ViT first obtains the output and gradient of each transformer block through a forward and backward propagation and then optimizes all transformer layers in a blockwise manner with a bottom-elimination second-order metric. And MPQ is straightforwardly embedded as a quantizer after Softmax function in the Matmul operation.

As for computation intensity, compared with existing methods, the execution process of APQ-ViT only needs to store the output and gradient of each transformer block instead of all the layers, which greatly reduces the storage footprint required for the entire process (reduced to about 20\%). It allows the process to be performed entirely in GPU memory to reduce the speed penalty caused by the data exchange with storage.

\begin{table}[t]
    \centering
    \captionsetup{width=1\linewidth}
    \caption{Ablation study of BBC and MPQ.}
    \label{tab:ablation}
    {
    \begin{tabular}{ccccccc}
    \toprule
    \textbf{\#bit(W/A)}  & \textbf{BBC} & \textbf{MPQ} & \textbf{ViT-S} & \textbf{ViT-B} & \textbf{DeiT-B} \\ \midrule
    Full-precision  &       &   & 81.39 & 84.54 & 81.80  \\ 
    \midrule
    \multirow{4}*{4/4}  &  &  & 42.57 & 30.69 & 64.39 \\ 
    ~  &  {\Checkmark} &   & 42.86 & 38.40 & {65.57} \\ 
    ~  &  & {\Checkmark}  & 46.16 & {35.44} & 66.98 \\ 
    ~  & {\Checkmark}  & {\Checkmark}  & \textbf{47.95} & \textbf{41.41} & \textbf{67.48} \\  
    \midrule 
    \multirow{4}*{6/6}  &    &    & 78.63  &  81.65 & 80.25  \\ 
    ~   &  {\Checkmark} &    & \dyf{78.78} & {81.84} & \dyf{80.33} \\ 
    ~   &  & {\Checkmark}  & 78.95 & 82.20  & 80.38\\ 
    ~   & {\Checkmark}  & {\Checkmark}  & \textbf{79.10} & \textbf{82.21} & \textbf{80.42}  \\  
    \midrule
       \multirow{4}*{8/8}  &    &  &  81.00 & 84.09  & 81.48 \\ 
    ~   &  {\Checkmark} &  & {81.01} & 84.18 & 81.63 \\ 
    ~   &  & {\Checkmark}  & 81.15 & {84.23} &  81.68 \\ 
    ~   & {\Checkmark}  & {\Checkmark}  & \textbf{81.25} & \textbf{84.26} & \textbf{81.72}  \\

     \bottomrule
    \end{tabular}
    }
\end{table}

\begin{table}[t]
    \centering
    \captionsetup{width=1\linewidth}
    \caption{Results of different post Softmax quantizers. }
    \label{tab:softmax-quant}
    \setlength{\tabcolsep}{1.9mm}
    {
    \begin{tabular}{ccccccc}
    \toprule
    \textbf{\#bit(W/A)}  & \textbf{Quantizer} & \textbf{ViT-S} & \textbf{DeiT-S} & \textbf{DeiT-B} \\ \midrule
    Full-precision  & - & 81.39 &  79.85 & 81.80  \\ \midrule
    
    \multirow{3}*{4/4}  & Log   &  {44.72}   &  21.59  &  60.91     \\ 
    ~ & Segmental & 37.70 & 22.31 & 60.02 \\
    ~ & MPQ & \textbf{47.95} & \textbf{43.55} & \textbf{67.48} \\    \midrule
    
    \multirow{3}*{6/6}  & Log   &  78.94  & 77.57  &  80.34     \\ 
    ~ & Segmental & 78.67 & 76.64  & 80.37 \\
    ~ & MPQ & \textbf{79.10} & \textbf{77.76} & \textbf{80.42} \\    \midrule

    \multirow{3}*{8/8}  & Log   & 81.13 & 79.76  &  81.70     \\ 
    ~ & Segmental & 81.00 & 79.47 & 81.70 \\
    ~ & MPQ & \textbf{81.25} & \textbf{79.78} & \textbf{81.72}\\  

     \bottomrule
    \end{tabular}
    }
\end{table}

\begin{table*}[!h]
    \centering
    \captionsetup{width=\linewidth}
    \caption{Comparison of different post-training quantization methods on image classification task with various vision transformer architectures and bit-widths.}
    \label{tab:classification}
    \setlength{\tabcolsep}{1.6mm}
    {
    \begin{tabular}{ccccccccccccc}
    \toprule
    \textbf{Method} &  \textbf{\#bit(W/A)}  & \textbf{ViT-T} & \textbf{ViT-S} & \textbf{ViT-S/32} & \textbf{ViT-B} & \textbf{DeiT-T} & \textbf{DeiT-S} & \textbf{DeiT-B} & \textbf{Swin-S} & \textbf{Swin-B} & \textbf{Swin-B/384} \\ \midrule
    Full-precision & 32/32    & 75.47 & 81.39 & 75.99 & 84.54 & 72.21  & 79.85  & 81.80  & 83.23  & 85.27  & 86.44  \\ \midrule
    
    FQ-ViT         & 4/4  & 0.10 & 0.10 & 0.10 & 0.10 & 0.10 & 0.10 & 0.10 & 0.10 & 0.10  & 0.10  \\
    % BasePTQ        & 4/4  & 0.10 & 0.12 & 0.71 & 0.16 & 0.13 & 0.13 & 0.19 & 1.81 & 0.30 & 0.25 \\
    PTQ4ViT        & 4/4  & 17.45 & 42.57 & 35.09 & 30.69 	& 36.96	& 34.08	& 64.39	& 76.09	&  74.02 & 78.84  \\
    APQ-ViT (Ours)          & 4/4  & \textbf{17.56} & \textbf{47.95} & \textbf{41.53} & \textbf{41.41}	& \textbf{47.94}	& \textbf{43.55} & \textbf{67.48} & \textbf{77.15}	&  \textbf{76.48}  & \textbf{80.84}\\ \midrule
    
    FQ-ViT  & 8/4  & 0.10 & 0.10 & 0.10 & 0.10 & 0.10 & 0.10 & 0.10 & 0.10 & 0.10  & 0.10 \\
    % BasePTQ & 8/4  & 0.13  & 0.10 & 2.60  & 0.10 & 0.39 & 0.43 & 0.38 & 8.98 & 8.55  & 3.34 \\
    PTQ4ViT & 8/4  & 36.17 & 63.00 & 49.68 & 71.64 & 48.00 & 36.08 & 70.07 & 80.13 & 81.45 & 83.75\\
    APQ-ViT (Ours)     & 8/4  & \textbf{38.62} & \textbf{67.17} & \textbf{64.57} & \textbf{72.47} & \textbf{56.28} & \textbf{41.31} & \textbf{71.69} & \textbf{80.62} & \textbf{82.08} & \textbf{83.87} \\\midrule
    
    FQ-ViT  & 4/8  & 27.84 & 71.17 & 43.93 & {78.48} & 64.42 & 74.70 & 79.19 & {81.17} & 81.43  & 82.69  \\
    % BasePTQ & 4/8  & 14.78 & 31.28 & 17.92 & 45.76 & 56.60 & 64.61 & 74.30  & 75.42 & 63.32 & 71.26 \\
    PTQ4ViT & 4/8  & 59.23 & 69.68 & 36.04 & 67.99 & 66.57	& 76.96	& 79.47	& 79.62	&  78.50  & 82.54   \\
    APQ-ViT (Ours)   & 4/8  & \textbf{59.42} & \textbf{72.30} & \textbf{61.81} & {72.63}	& \textbf{66.71} & \textbf{77.14}	& \textbf{79.55}	& {80.56}	&  \textbf{81.94} &  \textbf{83.42}    \\ \midrule
    
    FQ-ViT         & 6/6 & 0.38 & 4.26 & 2.65 & 0.10 & 58.66 & 45.51 & 64.63  & 66.50 & 52.09  & 0.10 \\
    % BasePTQ        & 6/6 & 51.48 & 69.79 & 59.57 & 76.96  & 67.11 & 72.57 & 78.80 & 81.75 & 83.42 & 85.26 \\
    PTQ4ViT        & 6/6 & 64.46 & 78.63 & 71.90 & 81.65   & 69.68 &  76.28  & 80.25 & 82.38	& 84.01 & 85.44 \\
    APQ-ViT (Ours)          & 6/6 & \textbf{69.55} & \textbf{79.10} & \textbf{72.89} & \textbf{82.21}	& \textbf{70.49} &  \textbf{77.76}	& \textbf{80.42}	& \textbf{82.67}	&  \textbf{84.18} &  \textbf{85.60}   \\ \midrule
    
    FQ-ViT         & 8/8 & 45.99 & 78.68 & 58.87 &  82.76 & 70.92 & 78.44 & 81.12  & 82.38 & 82.38 & 85.74 \\
    % BasePTQ        & 8/8 & 70.05 & 80.55 & 73.56 & 83.80 & 71.27 & 77.61 & 81.02 & 82.80  & 84.89  & 86.28 \\
    PTQ4ViT        & 8/8 & 74.56 & 81.00 & 75.58 & 84.25 & 71.72 & 79.47 & 81.48  & 83.10 & 85.14  & 86.36 \\
    APQ-ViT (Ours)           & 8/8 & \textbf{74.79} & \textbf{81.25} & \textbf{75.64} & \textbf{84.26} & \textbf{72.02} & \textbf{79.78} & \textbf{81.72}  & \textbf{83.16} & \textbf{85.16} & \textbf{86.40} \\

    % FQ-ViT         & 2/4  &  -   & 0.10 &  -   & 0.10 & 0.10 & 0.10 & 0.10 &     \\
    % PTQ4ViT        & 2/4  & 0.17 & 0.27 & 0.31 & 0.15 & 0.52 & 0.28 & 10.04 & 0.10 & 3.32 & 0.10      \\
    % Ours           & 2/4  & 0.15 & 0.33 & 0.27 & 0.16 & 0.52 & 0.21 & 7.69 & 0.10 & 2.50 & 0.13    \\ \hline
    
    % FQ-ViT         & 4/2 & - & 2.09 & - & 0.10 & 0.10 & 0.10 & 0.10 & 0.10       \\
    % PTQ4ViT        & 4/2 & 0.16 & 0.10 & 0.22 & 0.13 & 0.25 & 0.01 & 0.22 & 1.34 & 0.59 & 1.34      \\
    % Ours           & 4/2 & 0.17 & 0.18 & 0.33 & 0.14 & 1.26 & 1.26 & 0.14 & & 1.54 & 1.01    \\ 
    \bottomrule
    \end{tabular}
    }
\end{table*}

\section{Experiment}

In this section, we first demonstrate the fundamental pipeline of post-training quantization and the experimental settings. 
We start by ablation studies to evaluate the effectiveness of each proposed approach. And then We compare with other methods on both image classification and detection tasks with various vision transformer architectures. 

\subsection{Settings}

\textbf{PTQ scheme:} 
We first build a post-training quantization baseline for experiments, where we follow the \cite{yuan2021ptq4vit} to set the search range of weight and activation to $[0, 1.2]$ for image classification task and follow \cite{easyquant, liu2021post} and set to $[0.5, 1.2]$ for detection task, and evenly divide to $n=100$ intervals. he default search rounds of the alternative optimization is 3. We randomly select 32 images from the ImageNet dataset for classification tasks and only 1 image from COCO dataset for detection tasks.
We empirically set $\gamma=10$ as default. The bit-width of the quantized model is marked as W$w$A$a$ standing for $w$-bit weight and $a$-bit activation. As for comparison methods, we follow the official settings using the released codes. 

\textbf{Vision Tasks and Network Architectures:} 
To prove the versatility of our AFQ-ViT, we evaluate it on image classification and detection tasks. We adopt the most widely-used transformer-based networks for comparison, including ViT~\cite{dosovitskiy2020vit}, DeiT~\cite{Touvron2021deit} and Swin Transformer~\cite{ze2021swin} for classification task on ImageNet~\cite{Deng2009ImageNet} and three different scales of Swin Transformer for detection task on COCO dataset~\cite{lin2014coco}. 
Note that we do not quantize the activation in the first convolution layer and the last classification layer. We also keep the Softmax, LayerNorm and GeLU functions as full-precision since they cause little computational overheads, but quantizing them will cause a severe accuracy drop.

\begin{table*}[t]
    \centering
    \captionsetup{width=\linewidth}
    \caption{Comparison of different post-training quantization methods on object detection task under various bit-widths. }
    \label{tab:detection}
    \setlength{\tabcolsep}{1.6mm}
    {
    \begin{tabular}{cccccccccccc}
    \toprule
    \textbf{Method} &  \textbf{\#bit(W/A)} & \multicolumn{2}{c}{\makecell[c]{\textbf{Mask RCNN} \\ \textbf{Swin-T} \\ \begin{tabular}[c]{cc} AP$^{\mathrm{box}}$ & AP$^{\mathrm{mask}}$  \end{tabular}}}  & 
    \multicolumn{2}{c}{\makecell[c]{\textbf{Mask RCNN} \\ \textbf{Swin-S} \\ \begin{tabular}[c]{cc} AP$^{\mathrm{box}}$ & AP$^{\mathrm{mask}}$  \end{tabular}}}  & 
    \multicolumn{2}{c}{\makecell[c]{\textbf{Cascade Mask} \\ \textbf{RCNN Swin-T} \\ \begin{tabular}[c]{cc} AP$^{\mathrm{box}}$ & AP$^{\mathrm{mask}}$  \end{tabular}}} & 
    \multicolumn{2}{c}{\makecell[c]{\textbf{Cascade Mask} \\ \textbf{RCNN Swin-S} \\ \begin{tabular}[c]{cc} AP$^{\mathrm{box}}$ & AP$^{\mathrm{mask}}$  \end{tabular}}} & 
    \multicolumn{2}{c}{\makecell[c]{\textbf{Cascade Mask} \\ \textbf{RCNN Swin-B} \\ \begin{tabular}[c]{cc} AP$^{\mathrm{box}}$ & AP$^{\mathrm{mask}}$  \end{tabular}}} \\ \midrule

    Full-precision & 32/32  & 46.0 & 41.6 & 48.5 & 43.3 & 50.4 & 43.7  & 51.9 & 45.0 & 51.9 & 45.0  \\ \midrule
    
    % FQ-ViT    & 4/4   & 0.0 & 0.0 & 0.0  & 0.0   & 0.0 & 0.0 & 0.0 & 0.0  & 0.0 & 0.0 \\
    BasePTQ   & 4/4   & 0.9 & 0.9  & 12.6 & 11.8 & 1.3 & 1.2 & 8.4 & 7.7  & 4.0 & 3.7 \\
    PTQ4ViT   & 4/4   & 6.9 & 7.0 & 26.7 & 26.6 & 14.7 & 13.5 & 0.5 & 0.5 & 10.6 & 9.3 \\
    APQ-ViT (Ours)        & 4/4  & \textbf{23.7} & \textbf{22.6} & \textbf{44.7} & \textbf{40.1} & \textbf{27.2} & \textbf{24.4} & \textbf{47.7} & \textbf{41.1} & \textbf{47.6} & \textbf{41.5}  \\ \midrule
    
    % FQ-ViT     & 8/4  & 0.0 & 0.0 & 0.0 & 0.0  & 0.0   & 0.0 & 0.0 & 0.0  & 0.0  & 0.0 \\
    BasePTQ    & 8/4  & 2.2 & 2.1 & 23.2 & 21.4 & 3.2 & 3.0 & 16.3 & 14.6 & 7.6 & 6.6    \\
    PTQ4ViT    & 8/4   & 25.5 & 25.0 & 18.3 & 18.0 & 18.4 & 16.7 & 32.7 & 29.0 & 14.4 & 13.1   \\
    APQ-ViT (Ours)         & 8/4   & \textbf{33.7} & \textbf{31.6} & \textbf{46.6} & \textbf{41.8} & \textbf{36.6} & \textbf{32.2} & \textbf{49.6} & \textbf{43.4} & \textbf{49.2} & \textbf{43.0} \\ \midrule
    
    % BasePTQ         & 4/8   & 27.6 & 26.0 & 23.9 & 23.3 & 24.6 & 22.1 & 39.2 & 34.7 & 39.9 & 35.5      \\
    % FQ-ViT     & 4/8 & 19.9 & 18.8 & 34.8 & 32.3 & 29.0 & 25.5 & 41.9 & 36.7 41.2 & 36.1 \\
    BasePTQ & 4/8 & 42.4 & 38.8 & 45.7 & 41.0 & 45.5 & 40.0 & 47.4 & 41.5 & 47.2 & 41.6 \\
    PTQ4ViT  & 4/8   & 0.7 & 0.8  & 23.4 & 22.2 & 25.3 & 22.7 & 38.5 & 33.8 & 20.0 & 28.4   \\
    % Ours            & 4/8   & 43.0  & 39.4  & 42.1 & 39.1  & 46.5  & \dyf{40.9}  & \dyf{49.3} & \dyf{43.0} & \dyf{49.1} & \dyf{42.9}   \\ \midrule
    APQ-ViT (Ours)    & 4/8 & \textbf{43.1} & \textbf{39.3} & \textbf{47.3} & \textbf{42.2} & \textbf{46.2} & \textbf{40.7} & \textbf{49.4} & \textbf{43.0} & \textbf{49.0} & \textbf{42.9} \\ \midrule
    
     % BasePTQ     & 6/6   & 6.8  & 6.9  & 19.5 & 19.6 & 10.0 & 8.9  & 27.0 & 24.1 & 5.0  & 4.6 \\
    % FQ-ViT      & 6/6 & 41.8 & 38.5 & 38.1 & 34.8 & 46.5 & 40.6 & 39.9 & 35.0 & 30.6 & 26.9  \\
    BasePTQ     & 6/6  & 40.1 & 36.8 & 46.7 & 41.8 & 46.3 & 40.7 & 48.9 & 42.7 & 46.3 & 40.7 \\
    PTQ4ViT     & 6/6   & 5.8  & 6.8 & 6.5  & 6.6  & 14.7 & 13.6 & 12.5  & 10.8  & 14.2  & 12.9    \\
    APQ-ViT (Ours)              & 6/6   & \textbf{45.4} & \textbf{41.2} & \textbf{47.9} & \textbf{42.9} & \textbf{48.6} & \textbf{42.5} & \textbf{50.5} & \textbf{43.9} & \textbf{50.1} & \textbf{43.7} \\ \midrule
    
    % BasePTQ         & 8/8  & 21.2 & 20.6 & 8.5  & 8.9 & 18.1 & 16.1 & 17.7 & 15.8 & 22.8 & 20.4 \\
    FQ-ViT          & 8/8  & 45.3 & 41.2  & 48.2 & 42.6   & 49.7    & 43.3    & {51.7} & 44.2  & 51.1 & 44.3 \\
    BasePTQ    & 8/8 & 45.8 & 41.5 & 48.1 & 42.9 & 48.6 & 42.5 & 50.3 & 43.8 & 49.9 & 43.7 \\
    PTQ4ViT         & 8/8  & 28.0 & 27.1 & 1.5 & 1.4 & 40.3 & 35.6 & 20.8 & 18.7 & 2.0 & 1.9    \\
    % Ours            & 8/8  & \textbf{45.8} & \textbf{41.6} & \textbf{48.2} & \textbf{43.0}  & \textbf{48.9} & \textbf{42.7} & \textbf{50.7} & \textbf{44.1} & \textbf{50.2} & \textbf{43.9}      \\ \midrule
    APQ-ViT (Ours)    & 8/8 & \textbf{45.8} & \textbf{41.5} & \textbf{48.3} & \textbf{43.1} & \textbf{48.9} & \textbf{42.7} & \dyf{50.8} & \textbf{44.1} & \textbf{50.2} & \textbf{43.9} \\
    
    \bottomrule
    \end{tabular}
    }
\end{table*}
\subsection{Ablation Study}

We conduct extensive ablation studies on each proposed method. 
As Table~\ref{tab:ablation} shows, we evaluate our methods on ViT-S, ViT-B and DeiT-B vision transformer architectures. The baseline post-training quantization method suffers a severe accuracy loss, especially under 4-bit setting. While our methods retain the accuracy and the advantage becomes more obvious in lower bit-width. 
Applying the BBC can significantly improve the performance. For example, it helps ViT-B to get 38.40\% accuracy in W4A4 which is 7.71\% higher than the traditional method. 
As for the MPQ, we compare it with the Logarithmic quantizer and Segmental quantizer and evaluate ViT-S, DeiT-S and DeiT-B. As shown in Table~\ref{tab:softmax-quant}, MPQ quantizer outperforms other quantizers by a wide margin, especially in W4A4. DeiT-S equipped with MPQ is 21.96\% higher than the Logarithmic quantizer and 21.24\% higher than the Segmental quantizer. We conjecture that it is because in the lower bit-width condition, the quantizer spares fewer bits to represent the large magnitude, especially for Logarithmic and Segmental quantizers, the negative influence of damaging Matthew-effect becomes manifest. 
Besides, the phenomena are consistent in 6-/8-bit settings. 

Moreover, jointly applying the proposed methods can further improve the performance, which demonstrates that the BBC optimization strategy in conjunction with MPQ can work orthogonally.

\subsection{Comparison on Classification Task}

% We first compare our Information Distribution Guided Metric with several widespread metrics, including MSE, cosine distance, Hessian guided metric. We conduct comprehensive experiment on several different vision transformer architectures incluing ViT-S/B/L, DeiT-T/S/B and Swin-T/S/B.  As Table~\ref{tab:classification} shows, selecting the scale with the minimum information entropy in quantization error can lead to the optimal scale. 

We first conduct extensive experiments on ImageNet classification tasks. We choose different transformer-based architectures, including ViT, DeiT and Swin Transformer. The default patch size is 16$\times$16 and the image resolution is 224$\times$224 if not specifically mentioned (ViT-S/32 means the patch size is 32$\times$32, Swin-B/384 means the image resolution is 384$\times$384). 

% As the Table~\ref{tab:classification} shows, it achieves almost loss-less accuracy on 8-bit quantization. For instance, the DeiT-B and Swin-B/384 models quantized by APQ-ViT have only 0.08\% and 0.04\% accuracy drop, within 0.10\%. And the average accuracy loss of W8A8 compared with the full-precision model is only about 0.23\%. Moreover, under W4A8 setting, our method also outperforms previous methods by a clear margin. The average improvement compared with FQ-ViT is 5.05\%, and that with PTQ4ViT is 3.89\%. 

% We also evaluate APQ-ViT under even lower bit-width. 
% We observe that quantizing the parameters, especially activation, to 4-bit brings a severe accuracy drop. FQ-ViT almost crashes on the ViT series when the activation is lower than 8-bit, while our APQ-ViT recovers the accuracy significantly. 
% As shown in the table, the average accuracy improvement compared with PTQ4ViT under W6A6, W8A4 settings are 1.02\% and 3.87\% respectively. It is noteworthy that under W4A4 setting, our method shows an obvious advantage which achieves an impressive 5.07\% improvement on average. 

We highlight that APQ-ViT is versatile and has prevailing improvements over different transformer variants, patch sizes, input resolutions and bit-widths.
As Table~\ref{tab:classification} shows, our method shows an impressive advantage especially in lower bit-width (\ie W4A4). We observe that previous methods like FQ-ViT almost crash when quantizing the activations to lower than 8-bit, while our APQ-ViT improves the accuracy significantly by up to 5.17\% on average compared to PTQ4ViT under W4A4. For some specific models, like ViT-B, our method even outstrips PTQ4ViT by 10.72\%. 

Moreover, under W6A6 and W8A4 settings, the average accuracy improvement of APQ-ViT compared with the previous method is 1.02\% and 3.87\%, respectively. Under the W4A8 setting, our method accomplishes high accuracy. The average improvement compared with FQ-ViT is 5.05\%, and that with PTQ4ViT is 3.88\%. 
It is noteworthy that our methods achieve almost loss-less accuracy under W8A8. For instance, the DeiT-B and Swin-B/384 models quantized by APQ-ViT have only 0.08\% and 0.04\% accuracy drop, within 0.10\%. And the average accuracy loss of W8A8 compared with the full-precision model is only about 0.23\%. 

\subsection{Comparison on Object Detection Task}

To further evaluate the generalization capability of our methods, we extend it to object detection tasks using large-scale COCO datasets. 
We use the Mask RCNN and Cascade Mask RCNN detectors with Swin Transformers (Swin-T/S/B) as backbones. 

% The experimental results are presented in Table~\ref{tab:detection}. Compared to full-precision models, APQ-ViT only drops 0.18\% and 1.20\% under W8A8 setting with Mask RCNN detector and Cascade Mask RCNN detector. And APQ-ViT with W6A6, W4A8 settings also outstrips the BasePTQ by 3.25\% and 1.00\% on average with Mask RCNN, 2.28\% and 1.33\% on average with Cascade Mask RCNN. 

% We highlight that compared to classification tasks, quantizing activation to lower-bit, \eg W8A4 and W4A4, usually causes more damage to the model robustness and accuracy of detection tasks. Models calibrated by previous methods almost crash, while APQ-ViT recovers the accuracy by a large margin. Especially under W4A4 setting, the average improvement is a remarkable 24.43\% and 30.81\% over PTQ4ViT and BasePTQ respectively. For specific models, our APQ-ViT almost obtains comparable performance closer to full-precision counterparts. For example, our APQ-ViT achieves 44.7\% and 40.1\% (drops 3.8\% and 3.2\%) for AP$^\mathrm{box}$ and AP$^{\mathrm{mask}}$ with Cascade Mask RCNN Swin-S, while PTQ4ViT methods only get 26.7\% and 26.6\%. It shows great potential for low-bit quantized detectors to meet the accuracy requirements and be implemented in real-world applications. 

The results are presented in Table~\ref{tab:detection}.
We highlight that compared to classification tasks, quantizing activation to lower-bit, \eg W4A4 and W8A4, usually brings more challenges to model robustness and accuracy of detection tasks. Models calibrated by previous methods almost crash, while APQ-ViT converges and recovers the accuracy. Especially under the W4A4 setting, the average improvement is a remarkable 24.43\% and 30.81\% over PTQ4ViT and BasePTQ respectively. For example, our APQ-ViT achieves 44.7\% and 40.1\% (drops 3.8\% and 3.2\%) for AP$^\mathrm{box}$ and AP$^{\mathrm{mask}}$ with Cascade Mask RCNN Swin-S, while PTQ4ViT methods only get 26.7\% and 26.6\%. 

Furthermore, APQ-ViT with W6A6 and W4A8 settings also averagely outstrips the BasePTQ by 2.57\% and 1.20\%, respectively. As for W8A8, APQ-ViT gets comparable performance which only drops 0.18\% and 1.20\% compared to full-precision counterparts with Mask RCNN detector and Cascade Mask RCNN detector. It shows great potential for low-bit quantized detectors to meet the accuracy requirements and be implemented in real-world applications. 

In a nutshell, APQ-ViT is a versatile method that shows a great practical value on both image classification and object detection tasks over various bit-width and transformer variants. 

\section{Conclusion}
In this paper, we analyze the post-training quantization for vision transformers from optimization and structure perspectives and propose a novel method, namely APQ-ViT.
We first present a unified Bottom-elimination Blockwise Calibration scheme which fixes the overall quantization error in a blockwise manner and prioritizes the crucial errors that impact more on the final output. 
Moreover, we design a Matthew-effect Preserving Quantization to maintain the power-law distribution of Softmax and keep the function of the attention mechanism. 
Comprehensive experiments demonstrate that our APQ-ViT achieves prevailing improvements, especially in lower bit-width settings (\eg averagely up to 5.17\% improvement for classification and 24.43\% for detection on W4A4).
We highlight that APQ-ViT is a versatile method that works well with diverse vision transformer variants, including DeiT and Swin Transformer.

\section*{Acknowledgement}
This work was supported by The National Key Research and Development Plan of China (2021ZD0110503), National Natural Science Foundation of China (62022009 and 61872021) and Meituan. 

\clearpage
\bibliographystyle{ACM-Reference-Format}
\bibliography{sample-base}
\end{document}